# IDPL-PFOD2: A New Large-Scale Dataset for Printed Farsi Optical Character Recognition

Fatemeh Asadi-zeydabadi[1], Ali Afkari-Fahandari[1], Amin Faraji[1], Elham Shabaninia[2,*], Hossein Nezamabadi-pour[1]

[1] Intelligent Data Processing Laboratory (IDPL), Department of Electrical Engineering, Shahid Bahonar University of Kerman, Kerman, Iran.

[2] Department of Applied Mathematics, Faculty of Sciences and Modern Technologies, Graduate University of Advanced Technology, Kerman, Iran
eshabaninia@gmail.com

**Abstract**

Optical Character Recognition is a technique that converts document images into searchable and editable text, making it a valuable tool for processing scanned documents. While the Farsi language stands as a prominent and official language in Asia, efforts to develop efficient methods for recognizing Farsi printed text have been relatively limited. This is primarily attributed to the language's distinctive features, such as cursive form, the resemblance between certain alphabet characters, and the presence of numerous diacritics and dot placement. On the other hand, given the substantial training sample requirements of deep-based architectures for effective performance, the development of such datasets holds paramount significance. In light of these concerns, this paper aims to present a novel large-scale dataset, IDPL-PFOD2, tailored for Farsi printed text recognition. The dataset comprises 2,003,541 images featuring a wide variety of fonts, styles, and sizes. This dataset is an extension of the previously introduced IDPL-PFOD dataset, offering a substantial increase in both volume and diversity. Furthermore, the dataset's effectiveness is assessed through the utilization of both CRNN-based and Vision Transformer architectures. The CRNN-based model achieves a baseline accuracy rate of 78.49% and a normalized edit distance of 97.72%, while the Vision Transformer architecture attains an accuracy of 81.32% and a normalized edit distance of 98.74%.

# 1 INTRODUCTION

Optical Character Recognition (OCR) refers to the procedure of identifying and transforming words and characters from an image into editable and searchable text. The input images encompass a wide range of sources, such as PDF files, images taken with cell phones or cameras, and scanned documents [1]. OCR technology has a wide range of applications in different fields, including reading and processing forms and checks, converting physical copies of archived documents into digital formats, extracting written information like addresses from envelopes, enabling access to books and papers for individuals with visual impairments or reading difficulties, and more [2, 3]. An efficient OCR system facilitates these tasks by eliminating labor-intensive manual processes, resulting in saving time and effort.

Typically, an OCR system includes several main components, as depicted in Fig. 1. Firstly, the text to be recognized is obtained through various sources, like scanning physical documents, capturing images with a camera, or using any other imaging device. Next, the pre-processing step aims to improve the overall quality of the input image by implementing various operations such as noise reduction, skew correction, binarization, etc. It is noteworthy that effective pre-processing significantly enhances OCR performance. The pre-processed image is subsequently subjected to segmentation, which involves dividing the text lines into distinct parts such as characters, words, lines, and more. This step is contingent upon the specific recognition technique employed in the next stage. During the feature extraction step, the OCR system generates valuable characteristics from the input data and feeds them into the recognition module. These features are then utilized by the recognition module to identify and match the corresponding character based on the extracted characteristics. Next, the recognized text is passed to the optional post-processing module. It aims to refine and improve the accuracy of the recognized text, correct errors, and enhance the overall quality of the output [9]. Finally, the user receives the output.

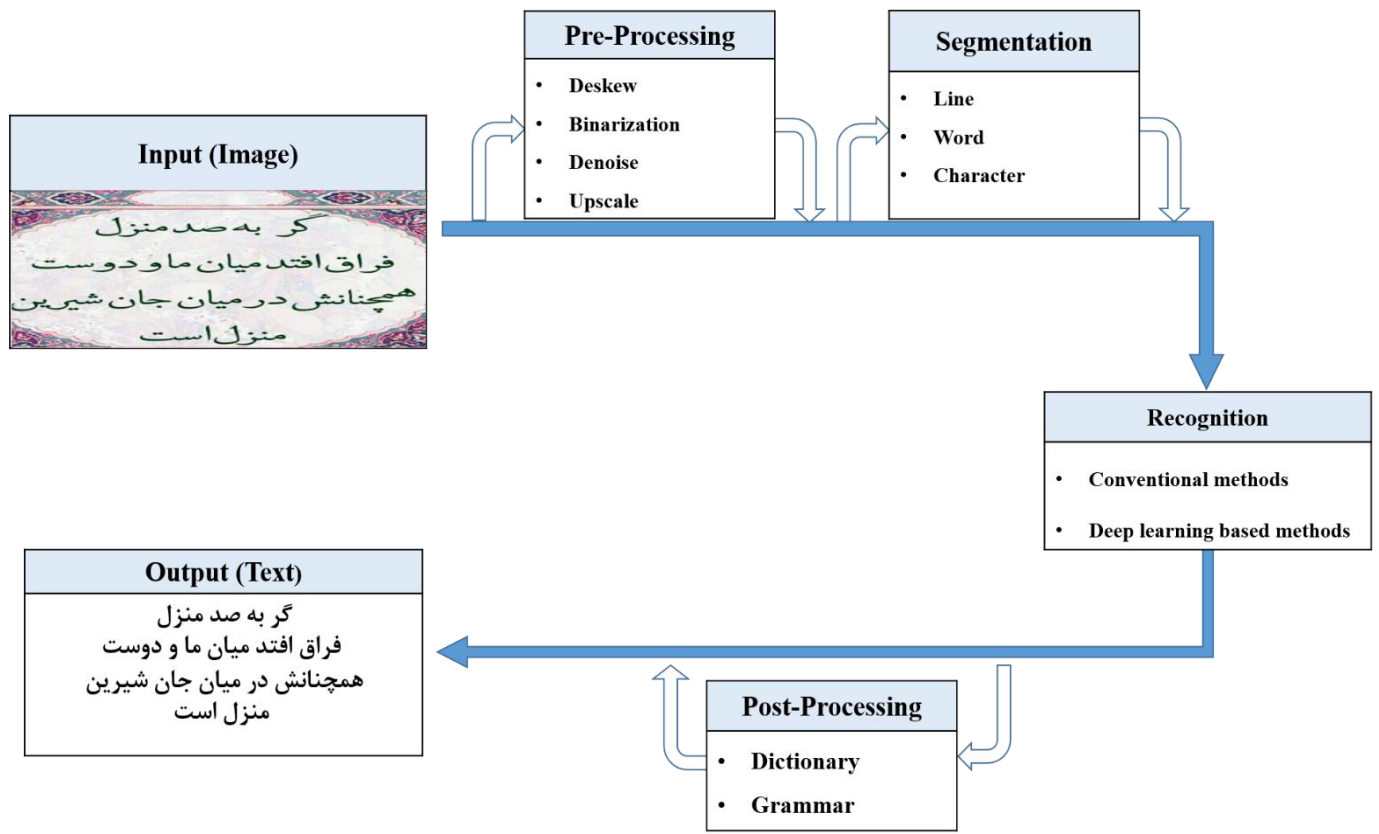

Fig.1 The overall steps of a general OCR system.

From a different perspective, OCR systems are classified in different ways; for instance, offline and online text recognition [4], where offline OCR systems work on available text images (like scanned documents and manuscripts) [5], and online systems recognize the text at the same time that it is recorded through a light stylus and a digitizer device [6, 7].

Basically, improving the quality of an OCR system involves factors like understanding the associated challenges in the chosen language, selecting an appropriate training algorithm, and providing a diverse and comprehensive dataset that covers different character styles in that language.

Farsi, also known as Persian, is a prominent Asian language spoken by over 150 million people worldwide [1]. It is a cursive right-to-left language and has 32 basic characters and 10 numerals (see Fig. 2), with certain characters having distinct writing modes, including prefix, postfix, and middle forms (see Fig. 3). While the Farsi language is distinct and maintains its unique vocabulary and grammatical structure, it extends over a spectrum of linguistic influences, intertwining with both the Arabic and Urdu alphabets. These similarities reveal a rich pattern of shared characters, writing conventions, and historical connections, creating an intricate linguistic landscape where Farsi bridges across cultural and regional boundaries [8]. Thus, considering the widespread use of Farsi and its significance in various domains, research on improving technologies like OCR holds the potential to benefit a broad audience. Moreover, by developing accurate OCR systems tailored for Farsi, we can enhance accessibility, information retrieval, and language processing capabilities for Farsi-speaking communities worldwide. However, developing an accurate OCR system for Farsi presents significant challenges due to the unique characteristics of the language. Below are several of the primary challenges associated with the Farsi language [3]:

1. An example word in Farsi, while certain characters are connected, others remain separate in Farsi script (see Fig.3.a).

2. The shape of characters may vary depending on their position within the word (see Fig.3.b).

3. certain characters may exhibit overlapping as they are written. For instance, consider the two characters "ن" and "گ" in the word "انگور" (see Fig.3.c).

4. In Farsi, words are constructed by connecting contiguous characters that overlap with one another, making it challenging to segment them into distinct characters. For instance, in the word "تلاش", segmenting "ت" and "ل" or "ل" and "ا" in "لا" is challenging, since they cannot be easily segmented by a straightforward vertical line (see Fig.3.d).

5. The distinction among certain Farsi characters, such as "ث", "ت", "غ" and "ع" differ from each other in terms of the dot(s) placement. As a result, the OCR system may encounter difficulties when attempting to recognize these characters. In the Farsi language, 18 characters are distinct from each other based on the dot(s) placement (see Fig.3.e).

6. Stretching of the characters; for example "ت" in the word "تا" (see Fig.3.f).

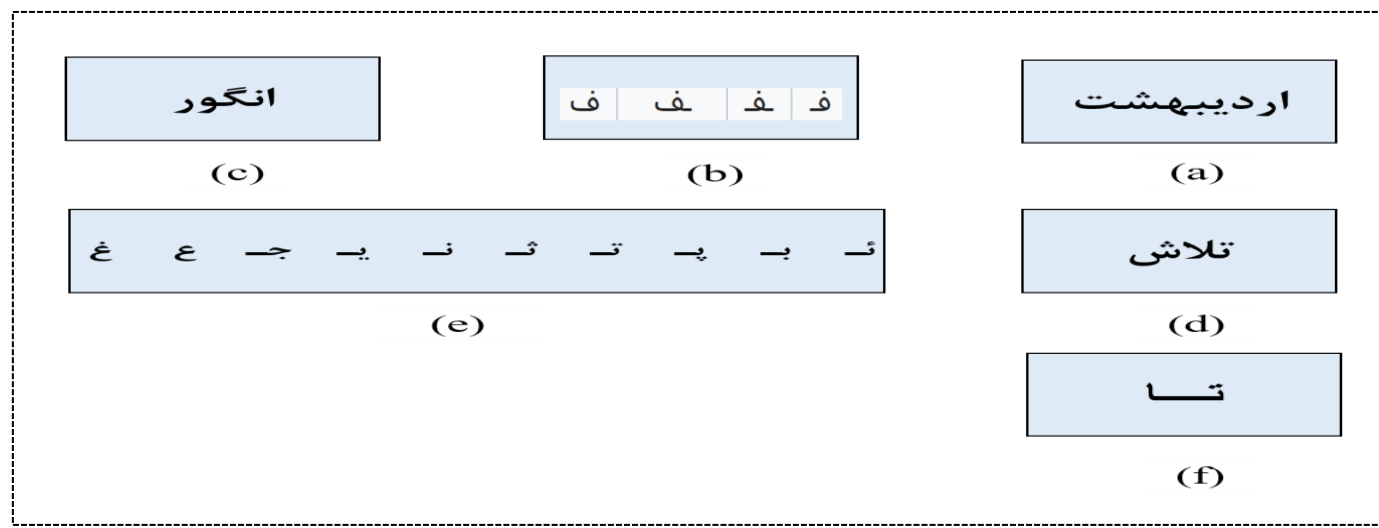

Fig.2 Farsi Character and Numeral Set.

Fig.3 Some general features and challenges of the Farsi script, which is also valid for Arabic and Urdu scripts [2].

Despite the long history of OCR research, ongoing scientific research is still being conducted in this field. That is due to the existence of various challenges, such as texts in multiple languages or with different font sizes, different perspectives and inappropriate lighting in camera photography, blurred images, crowded or dark backgrounds, and more. In the context of Farsi-like scripts, the continuity of characters and the great similarity between characters (some characters differ only in one dot) also add to these challenges and make it difficult to recognize texts.

As mentioned above, Farsi is significantly different from Latin languages. Although there is a great number of research on OCR in these languages, there are limited works on Farsi OCR. Additionally, the scarcity of OCR datasets for the Farsi language represents a noticeable challenge in the research and development of accurate OCR systems. Without a sufficient amount of diverse and high-quality training data, OCR systems may struggle to effectively recognize and process Farsi text, which is further complicated by the language's unique script and characteristics. These limitations underscore the need for the creation and expansion of comprehensive Farsi datasets to facilitate the accuracy and performance of OCR systems tailored to the Farsi language. To summarize, the key contributions of this paper are as follows:

1. In this paper, IDPL-PFOD2, an innovative large-scale dataset for Farsi OCR, is presented. To the best of our knowledge, this is the first publicly available large-scale dataset, comprising over 2 million images that encompass a wide range of fonts, sizes, and styles for characters, words, and sentences.

2. Beyond meeting the demands of the Farsi language, the purposed dataset, with its comprehensive coverage of Farsi's distinctive cursive script and various writing modes, serves as a valuable asset for researchers endeavoring to develop OCR systems. Furthermore, it can play a prominent complementary role in developing OCR systems for languages that share linguistic affiliations with Farsi, like Arabic and Urdu.

3. Finally, to evaluate the performance and effectiveness of the proposed dataset, two deep-based architectures, namely a CRNN-based and a Transformer-based model, were employed.

The remaining sections of this paper are organized as follows: Section 2 reviews related work and existing datasets. The details of the newly generated dataset are discussed in Section 3. In section 4, we delve into experiments on the proposed dataset. Finally, section 5 concludes the paper.

## 2 Related work

In this section, a comprehensive overview of recent developments in OCR technology is presented. To facilitate a better understanding of these developments, the review commences by exploring deep-based OCR methods for both English and Farsi-like languages in sections 2.1 and 2.2, respectively. Subsequently, in section 2.3, recent Farsi OCR datasets are discussed.

### 2.1 Deep-based OCR methods in the English language

The field of optical character recognition (OCR) has made substantial progress, primarily owing to the rise of deep learning techniques and advances in hardware capabilities. Hence, the research community has dedicated a considerable amount of effort to developing OCR techniques. For a more extensive review of state-of-the-art OCR methodologies, readers are encouraged to refer to the survey papers provided in [9-12]. In the early stages of OCR research, convolutional networks found their particular utilization in recognizing discretely written characters. These networks, known for their effectiveness in classifying predefined categories or classes, were primarily used for tasks involving the recognition of numbers and individual characters [13]. For instance, in [14], after segmentation, characters are identified by a CNN that was previously trained by labeled data. Similarly, in [15], the task of scene-text recognition is addressed as an image classification problem, where the objective is to categorize 90,000 words from the English dictionary.

However, as OCR technology advanced, there emerged a need for more versatile models capable of handling complex and context-rich text recognition, leading to the evolution of deep-based architectures laid on the foundation of convolutional recurrent neural networks (CRNNs) and Transformers [16-19]. These architectures expanded the horizons of OCR, enabling the recognition of full words or sentences with improved accuracy and efficiency through various scenarios. For example, significant advancements in OCR technology have been achieved, with a particular focus on sequential architectures like CRNN-based networks, which are comprised of convolutional layers followed by a recursive network. In this architecture, convolutional layers play a pivotal role in extracting crucial features from input images, enhancing the system's robustness against local distortions. Consequently, these networks have been widely adopted as feature-extraction modules [20, 21]. On the other hand, recursive networks have demonstrated their effectiveness in sequence modeling, contributing to the overall accuracy of OCR systems, which inherently involve image-based sequence prediction. The essence of sequence modeling lies in bridging the gap between visual features and text recognition. It records textual information as a sequence of characters for the next step to recognize each character. Moreover, to address the challenge of character recognition at each time step, particularly in the absence of prior knowledge regarding the alignment between image pixels and target characters, recent innovations have introduced methods based on connectionist temporal classification (CTC) and attention mechanisms [22-24]. More specifically, a group of studies investigated the utilization of the CTC function after RNN layers to align the networks' predictions with the actual textual information in the image. For instance, the work done in [20], introduces a CRNN-based architecture for scene-text recognition. It allows end-to-end training, accommodates sequences of different lengths without requiring character-level segmentation or scale normalization as well as being versatile with and without predefined lexicons.

Recently, Transformer-based architectures [25] have gained popularity in OCR tasks due to their ability to capture long-range dependencies and contextual information. These models leverage the self-attention mechanism, allowing them to dynamically attend to more relevant portions of the sequence. One key difference from CRNN-based networks is that Transformers do not rely on sequential processing, making them inherently parallelizable. This feature can lead to faster training and inference times, joined with improved efficiency in various scenarios. For instance, the works done in [26-28] are among the most recent studies that deployed this structure to effectively address the OCR scenarios.

## 2.2 Deep OCR methods for Farsi-like languages

This section provides a focused examination of deep OCR methods applied in Farsi and related languages like Arabic and Urdu. These languages share the characteristic of right-to-left script and possess similar alphabets, although they differ in the number of characters. However, due to similarities between these languages, the techniques used to solve OCR problems in one language may be adapted for other Farsi-like languages as well. In reference [29], a model for Arabic OCR is presented. This model incorporates BLSTM networks in conjunction with CTC to accurately predict relevant Arabic text sequences. Moreover. An additional linguistic model has been integrated into the model to enhance the accuracy of the output at the prediction stage. In [30], the trained weights of AlexNet and GoogleNet models are utilized for the classification of 54 letters and handwritten numbers in Urdu. In [31], discrete Arabic character features are extracted through a convolutional structure, followed by classification using a support vector machine in the final layer. A similar approach is also presented in [32] for the identification of Farsi handwritten characters. In [33], a convolutional neural network is proposed for the identification of Urdu ligatures. In [34], the authors investigated the accuracy of classifying colored numbers in Farsi manuscripts. They employed a convolutional network along with a dataset comprising 13,330 color images of the 0-9 digits. Additionally, the work done in [35] employs a multi-dimensional bidirectional LSTM (MD-BLSTM) coupled with a CTC approach to perform text recognition on Arabic-printed documents while eliminating the requirement for character-level segmentation. Likewise, in [36], an architecture combining CNNs with BLSTM and CTC is employed for the recognition of Arabic handwritten texts.

Furthermore, the work described in [37] utilizes a similar structure for recognizing Arabic scene texts. In addition, recent research works have successfully employed Transformer-based architectures for Arabic handwritten and printed text recognition [38-40]. For instance, a Transformer-based model tailored for recognizing Arabic historical handwritten text is introduced in [40]. The authors highlight challenges posed by the variability and absence of a standardized style in Arabic handwriting. Also, the OCR model integrates ResNet101 and Transformer to capture spatial and temporal dependencies, with emphasis on the latter for enhanced contextual information and improved accuracy. On the other hand, similar advancements in the field have applied the Transformer-based architecture to the domain of Urdu handwritten [41, 42]. Table 1 summarizes the recent works on Farsi, Arabic, and Urdu languages using deep-based methodologies.

Table 1 A summary of deep-based methodologies in Farsi, Arabic, and Urdu.

| Ref | Language | Model | Year | Dataset | Evaluation criterion | Performance |
|---|---|---|---|---|---|---|
| [43] | Urdu | MLSTM | 2016 | VPTI | Accuracy | 98% |
| [44] | Arabic | TDNN | 2016 | OIHACDB-28/ OIHACDB-40/ AIA9K | Accuracy | 94.7%, 98.3%, 95.6% |
| [49] | Arabic | CNN-SVM | 2016 | HACDB/ IFN/ ENIT | Accuracy | 94.17%, 92.95 |
| [45] | Arabic | CNN | 2017 | private dataset | Accuracy | 94.90% |
| [46] | Arabic | CNN | 2017 | HD/HPD /HSD | Accuracy | 81%, 85%, 83.5% |
| [47] | Arabic | CNN(VGG) | 2017 | HACDB | Accuracy | 99.57% |
| [48] | Arabic | CNN | 2017 | AIA9K/ AHCD | Accuracy | 94.80%, 97.60% |
| [32] | Persian | CNN(LeNet-5) | 2017 | Hoda | Accuracy | 97.10% |
| [36] | Arabic | CNN-BLSTM-CTC | 2018 | IFN/ ENIT | Accuracy | 92.21% |
| [49] | Persian | CNN-SVM | 2018 | private dataset | Accuracy | 61.14% |
| [50] | Urdu | LSTM-RNN | 2019 | CLETI/ VPTI | Accuracy | 99.00% |
| [48] | Urdu | CNN(AlexNet) CNN(GoogleNet) | 2020 | IFHCDB | Accuracy | 96.30%, 94.7% |
| [42] | Urdu | CNN-Transformer | 2022 | NUST-UHWR | CER | 5.31 |
| [41] | Urdu | Vision Transformer | 2023 | UNHD/ NUST-UHWR/ UHLD | Accuracy | 97.2%, 97.6%, 95.3 |
| [39] | Arabic | Transformer | 2023 | KAFD/SANAD | CER, WER | 0.8, 2.3 |
| [38] | Arabic | Transformer | 2023 | KHATT | CER | 19.76 |

## 2.3 Farsi OCR datasets

As discussed earlier, to effectively train OCR systems utilizing deep-based techniques, it is essential to have access to extensive and diverse image datasets with substantial quantities. However, a notable gap exists in the availability of publicly accessible datasets containing a significant number of samples for the recognition of printed Farsi texts. Table 2 shows existing datasets for Farsi OCR. Based on the information provided in this table, it is evident that only three image datasets for Farsi printed texts are accessible to the public, such as AUT-PFT [51], Shotor[1] [52], and IDPL-PFOD [53]. In the following, these datasets are briefly discussed.

Table 2 Existing Farsi datasets.

| Dataset | Type of content | Samples | Type of samples | Availability | Year | Reference |
|---|---|---|---|---|---|---|
| CENPARMI | Date / Digit (Persian)/ The string of digits/ character/ Word (in bank checks)/ Digit (English, written by Persian speakers) | 175/ 18000/ 7350/ 11900/ 8575/ 3500 | Handwritten | Public | 2006 | [54] |
| IFHCDB | Character/ Digit | 52380/ 17740 | Handwritten | Public | 2006 | [55] |
| Hoda | digit/ Character | 102352/ 88351 | Handwritten | Public | 2007 | [56] |
| FHT | Sentence / Word / Sub-word | 8050/ 106600/ 230175 | Handwritten | Public | 2009 | [57] |
| AUT-PFT | Word | 10000 | Printed | Public | 2015 | [51] |
| Sadri | Date (in numbers)/ Date (in character)/ Digit (Persian)/ string of digits Character/ Symbols and punctuation/ Mathematical symbols/ The most used Persian words in the field of names of people, products | 2500/ 2000/ 97124/ 450/ 43000/ 11500/ 16000/ 70000 | Handwritten | Public | 2016 | [59] |
| Shotor | Word | 120000 | Printed | Public | 2020 | [52] |
| IDPL-PFOD | Line (part of sentences) | 30138 | Printed | Public | 2021 | [53] |

---

[1] https://github.com/amirabbasasadi/Shotor

| PECI | Text line | 2000000 | Printed | Private | 2022 | [61] |
| IDPL-PFOD2 | Phrase | 2003541 | Printed | Public | 2023 | Ours |

**AUT-PFT [51]:** There are 10,000 word-level images in this dataset, which was published in 2015 and contains 127 unique characters. Notably, the characters are uniformly distributed, resulting in words without specific meaning. Moreover, the dataset includes images that simulate printed and scanned text with realistic noise. The examples in the dataset have 10 commonly used Farsi fonts and 4 different font sizes. The dataset also provides ground-truth data in XML file format.

**Shotor [52]:** Published in 2020 and contains 120,000 grayscale 50*100 images with their corresponding words. Each of the dataset samples has a meaningful Farsi word written in different fonts and sizes. Additionally, the meaningful words in this dataset are extracted from Farsi Wikipedia[2] and Ganjoor Website[3].

**IDPL-PFOD [53]:** Published in 2021, it uses two Farsi text corpora: Miras and Hamshahri. The length of sentences in this dataset includes 15 words. As a result, a total of 30,138 text-line images, equivalent to 452,070 words with a dimension of 50*700 pixels, have been generated. In this work, a diverse range of fonts and random font sizes varying from 10 to 16 have been applied. Besides, three types of backgrounds, such as plain white, noisy, and textured, are utilized. Furthermore, to enhance the realism of the dataset, distortion and blur effects are incorporated into the images.

---

[2] https://fa.wikipedia.or
[3] https://ganjoor.net

# 3 BUILDING THE IDPL-PFOD2 DATASET

The following subsections provide an overview of the process employed to generate the IDPL-PFOD2 dataset. This process involved the creation of synthetic text and the transformation of that text into phrase-level text images. Fig.4 provides an overview of the process employed in generating the IDPL-PFOD2 dataset.

## 3.1 Data collection

The aim of the initial phase of the dataset generation procedure is to generate a Farsi text corpora suitable for creating an image dataset. As previously stated, our goal is to construct a single-level corpus comprised of text phrases and cohesive word sequences that convey meaningful information, whether they form complete sentences or not. Moreover, through the thoughtful selection of text phrases, the resulting textual image is ensured to accurately represent the diversity and complexity of real-world Farsi text such as variation in fonts, sizes, and styles, making it well-suited for the development of efficient OCR systems. Consequently, a diverse array of Farsi-language PDF files, comprising academic dissertations from a wide spectrum of disciplines was chosen.

## 3.2 Data Cleaning

To achieve an accurate and consistent text, the initial raw text is subjected to a series of data-cleaning steps, such as removing irrelevant and undefined characters or symbols, handling special cases like misspellings, and standardizing the text's format. Moreover, during the data cleaning step, removing irrelevant and undefined characters or symbols may introduce so-called "space" or "whitespace" characters in the text. These characters refer to any characters, symbols, or sequences that are used to represent empty spaces within a text. The most common type of space character is a simple space, which is a visually empty character. However, there are other types of space characters as well, such as tabs, line breaks, and various non-breaking spaces. Eventually, along with various data cleaning steps, these characters are removed and clean text is generated. On the other hand, the processed text is often subjected to excessively long text lines. So, to produce consistent and uniform text samples throughout the dataset, a standard length of 25 words is established to tackle this issue as well. According to this, a total count of 2,003,541 phrases was generated as the final output of this step.

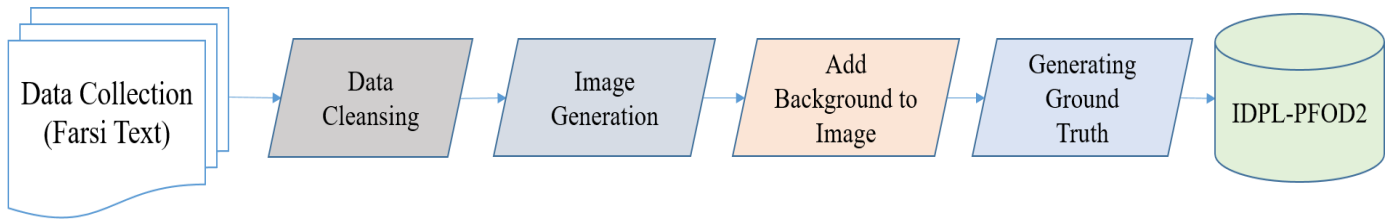

Fig.4 Dataset generation process.

## 3.3 Image Generation

Defining the fonts, font sizes, and font styles is a critical preliminary step before generating phrase-level images. In the case of Farsi language text, numerous fonts with various styles are accessible. However, not all of them are commonly used, with many primarily designed for graphic purposes. In the context of the IDPL-PFOD2 dataset, based on the guidelines provided by the Iran Supreme Council of Information and Communication Technology (ISCICT), a choice was made to utilize 11 fonts, including Badr, Zar, Roya, Nazanin, etc. These fonts were thoughtfully chosen for their compatibility with machine learning and data-driven applications, as well as their strict adherence to standardized formats. Moreover, to enhance the diversity, we also utilized two styles for most fonts, such as "Bold" and "Normal". This approach is consistent with recent research that has shown the benefits of incorporating multiple styles per font to improve the performance of machine learning models. Additionally, according to recent studies, the inclusion of various font sizes within a dataset has been shown to offer advantages in constructing a more authentic and diverse portrayal of the data. Therefore, a random font size in the range of 16 to 20 has been selected for the fonts.

Once the fonts, font sizes, and styles have been specified, image generation may initiate. The process begins by rendering the text using the designated font and formatting parameters to produce an image containing the specified phrase-level text. This resulting textual image is subsequently overlaid onto an image background, a plain canvas, or a textured image. The entire image generation process was facilitated using the Python programming language, and the resulting images were saved in "PNG" format. Each collection of images associated with a specific font was organized into distinct folders named after the respective font. Furthermore, the dimensions of the generated images were uniformly set at 300*50 pixels.

### 3.4. Add Background to image

The creation of the IDPL-PFOD2 dataset was explicitly aimed at facilitating the development of Farsi OCR systems. To promote this objective, a diverse array of backgrounds has been integrated into the generated images, providing researchers with the flexibility to employ the dataset for both printed and scene-text recognition scenarios. Notably, we incorporated three distinct background types, including plain white and pink, noisy, and textured backgrounds. The noisy background category involved the introduction of salt and pepper, gaussian, and speckle noises to simulate real-world environmental noise. As shown in Fig.5, in the case of textured backgrounds, a total of 12 unique patterns were employed to produce visually engaging and diverse background variations. Besides, to ensure the resulting dataset maintains a sense of balance and accurately represents each font, we carefully generated a nearly equal number of images for every font style. In particular, we employed three distinct background types for the images within each font category, distributing them evenly among plain white and pink, noisy, and textured backgrounds.

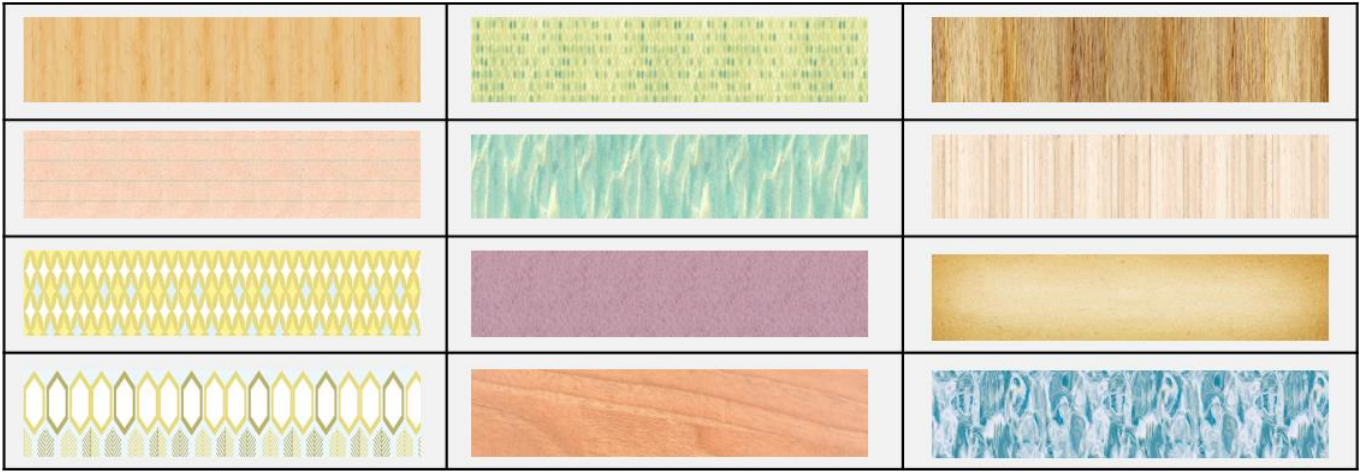

Fig.5 12 types of textured backgrounds used to embed textual contents.

For fonts like Zar, Badr, Nazanin, and Yekan, a total of 67,931 images were crafted for each of the three background variations. As for the remaining fonts, the total count of generated images for each background type was set at 37,931. This approach ensures that the dataset exhibits a harmonious distribution of fonts and backgrounds for comprehensive evaluation and analysis. On the other hand, when considering font styles (i.e., normal and bold), the dataset encompasses a total of 537,241 images for the normal style and 130,606 images for the bold style across the various background variations. Furthermore, considering the text line samples, there are a total of 36,031 distinct phrases, while at the word level, there are 25,315 unique words. Table 3 summarizes the dataset's statistics.

### 3.5 Generating Ground truth

In the final phase of creating the IDPL-PFOD2 dataset, ground-truth information is provided for each image. This process involves the task of establishing accurate and reliable reference data for textual content found within the images and is of paramount importance for training and assessing OCR models. Hence, a TXT file containing the ground truth is considered to offer both human-readable and machine-readable formats, ensuring its accessibility and ease of use for fellow researchers engaging with the dataset. This file comprises two columns, where the first column provides the image names and the second column contains the corresponding image's textual information. Moreover, the dataset requires approximately 19.24 gigabytes of storage space and is conveniently accessible[1].

Table 3 IDPL-PFOD2 dataset statistics.

| Font | Backgrounds | | | | | |
|---|---|---|---|---|---|---|
| | Style | | | | | |
| | Normal | | | Bold | | |
| | Plain | Noisy | Texture | Plain | Noisy | Texture |
| Badr | 67931 | 67931 | 67931 | 7175 | 7175 | 7175 |
| Zar | 67931 | 67931 | 67931 | 7175 | 7175 | 7175 |
| Yougut | 37931 | 37931 | 37931 | 7175 | 7175 | 7175 |
| Roya | 37931 | 37931 | 37931 | 7175 | 7175 | 7175 |
| Traffic | 37931 | 37931 | 37931 | 7175 | 7175 | 7175 |
| Titr | 37931 | 37931 | 37931 | 7175 | 7175 | 7175 |
| Mitra | 37931 | 37931 | 37931 | 7175 | 7175 | 7175 |
| Lotus | 37931 | 37931 | 37931 | 7175 | 7175 | 7175 |
| Compset | 37931 | 37931 | 37931 | 7175 | 7175 | 7175 |
| Nazanin | 67931 | 67931 | 67931 | 66031 | 66031 | 66031 |
| Yekan | 67931 | 67931 | 67931 | - | - | - |
| Total lines | 537241 | 537241 | 537241 | 130606 | 130606 | 130606 |
| | 1611723 | | | 391818 | | |
| | 2003541 | | | | | |
| Unique Words | 25315 | | | | | |
| Unique Phrases | 36031 | | | | | |
| Dimensions | 300*50 | | | | | |

As shown in Fig. 6, a visual depiction of 15 different sample images from the dataset is presented, while Fig.7 outlines the specification of images throughout the dataset. On the other hand, according to Table 4, a comparative analysis is provided to showcase a thorough examination of the features of the proposed dataset in contrast to other publicly available Farsi printed text datasets. This comparison is based on the diversity and quantity of available samples, offering valuable insights into the dataset's unique attributes and advantages.

---

[1] https://github.com/ftmasadi/IDPL-PFOD2

| Font | Style | Texture | Noise | Plain |
|------|-------|---------|-------|-------|
| Lotus | Bold | وزن‌های سیناپـسی برای | درنظرگرفتن اسکریپت‌های غیر | تصاویر مورداستفاده اسنادی |
| Badr | Normal | به سبب سر و وضع و طرز لباسش | و بازی های جالب هم انتخاب کرد | ر این کتاب شما می توانید بین تعدادی زیادی از |
| Titr | Bold | تفسیر داده‌ها مستقل | متن‌های لاتین چاپی | از روش‌های مختلف |
| Zar | Normal | در سیاره شازده کوچولو دانه های | به محث سوختن گوگرد در کتاب فـسر به | ملکی که پریشان شد از شومی |
| Nazanin | Bold | حشرات بدون بال نمی توانند بروار | چون تمام سیاره را فرا می گیرد | بر هیچ دلی مبـاد و بر هیچ |

Fig. 6 Fifteen different images from the IDPL-PFOD2 dataset with various backgrounds, fonts, and styles.

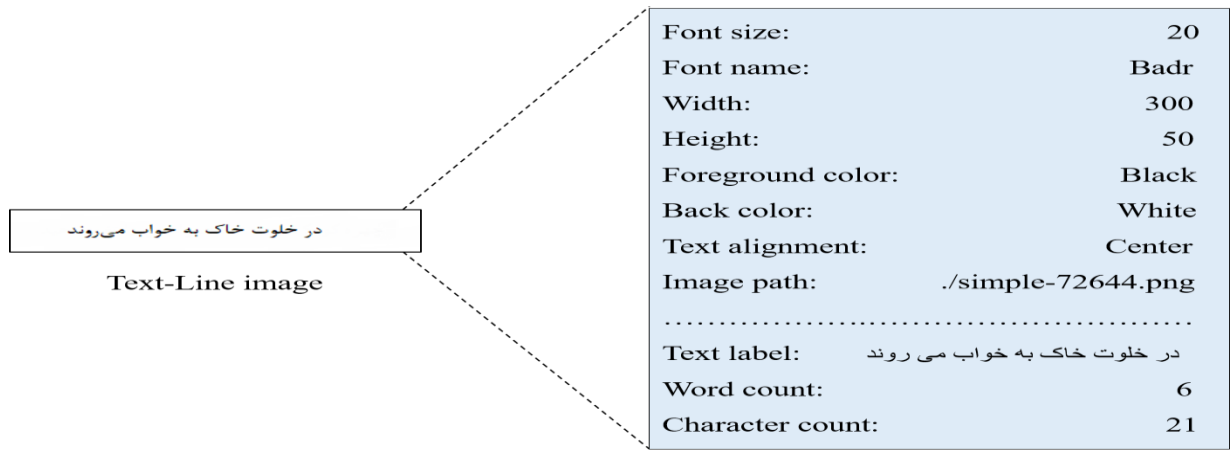

Fig. 7 Information details of a sample image from the IDPL-PFOD2 dataset.

Table 4 The comparison of publicly available datasets with IDPL-PFOD2 for printed Farsi OCR.

| Dataset | Samples | Type of samples | Plain | Noisy | Texture | Year |
|---------|---------|-----------------|-------|-------|---------|------|
| **AUT-PFT** | 10000 | Meaningless words | ✓ | ✓ | - | 2015 |
| **Shotor** | 120000 | Words | ✓ | - | - | 2020 |
| **IDPL-PFOD** | 30138 | Lines | ✓ | ✓ | ✓ | 2021 |
| **IDPL-PFOD2** | 2003541 | Phrase | ✓ | ✓ | ✓ | 2023 |

# 4 EXPERIMENTS

In this section, the authors offer a more comprehensive overview of the results obtained from various methods applied to the IDPL-PFOD2 dataset. To establish a solid foundation for future research, the evaluated models are introduced in Section 4.1. Subsequently, the evaluation metrics and implementation details are discussed in Sections 4.2 and 4.3, respectively. Finally, in Section 4.4, a detailed analysis of the achieved results is presented.

## 4.1 MODELS

As mentioned earlier, to assess the efficacy of the IDPL-PFOD2 dataset for developing OCR systems, two state-of-the-art deep-based architectures have been employed. These architectures include a CRNN-based and a Transformer-based (vision Transformer) network [20, 62]. The following subsections provide an in-depth review of these models.

### 4.1.1 Convolutional recurrent neural network (CRNN-based) architecture

Over the last decade, the remarkable performance of deep-based techniques, coupled with advancements in hardware processing power, has enabled researchers to address a multitude of complex tasks. Consequently, these techniques have found successful application across diverse domains of OCR scenarios, such as printed text, handwritten documents, and scene-text recognition. Among the most prevalent deep-based models are those based on the combination of convolutional neural networks (CNNs), recurrent networks, and connectionist temporal classification (CTC) layer [63]. This architecture is particularly well suited for tasks that involve sequential data, such as the recognition of a string of characters within a desired image. The overall architecture of a CRNN-based network for character recognition is depicted in Fig.8, which includes three distinct stages, the convolutional layer, recurrent layer, and transcription layer. Firstly, input images undergo a series of convolutional layers (MobileNetV3), which are responsible for extracting feature sequences. These feature sequences serve as the input for the subsequent stage. In the next stage, a deep bidirectional long short-term memory (BiLSTM) network is employed. With bidirectional LSTMs, one set of LSTM units processes the input sequence from left to right, while the other processes it from right to left. This bidirectional approach allows the model to capture contextual information in both directions, making it a valuable feature for OCR, where character recognition often hinges on the characters preceding and following each symbol.

Finally, in the transcription layer, a connectionist temporal classification (CTC) layer is introduced after the bidirectional LSTMs. CTC is particularly beneficial for sequence-to-sequence tasks, offering alignment-free flexibility. This means the OCR model can recognize text of varying lengths and irregular layouts without requiring prior knowledge of the alignment between input images and their corresponding text. Throughout the OCR process, at each time step, the CTC layer takes the per-frame predictions produced by the recurrent layers as input and generates a probability distribution encompassing all conceivable label sequences that could have produced the input sequence. Ultimately, the label sequence with the highest probability emerges as the definitive output of the OCR system.

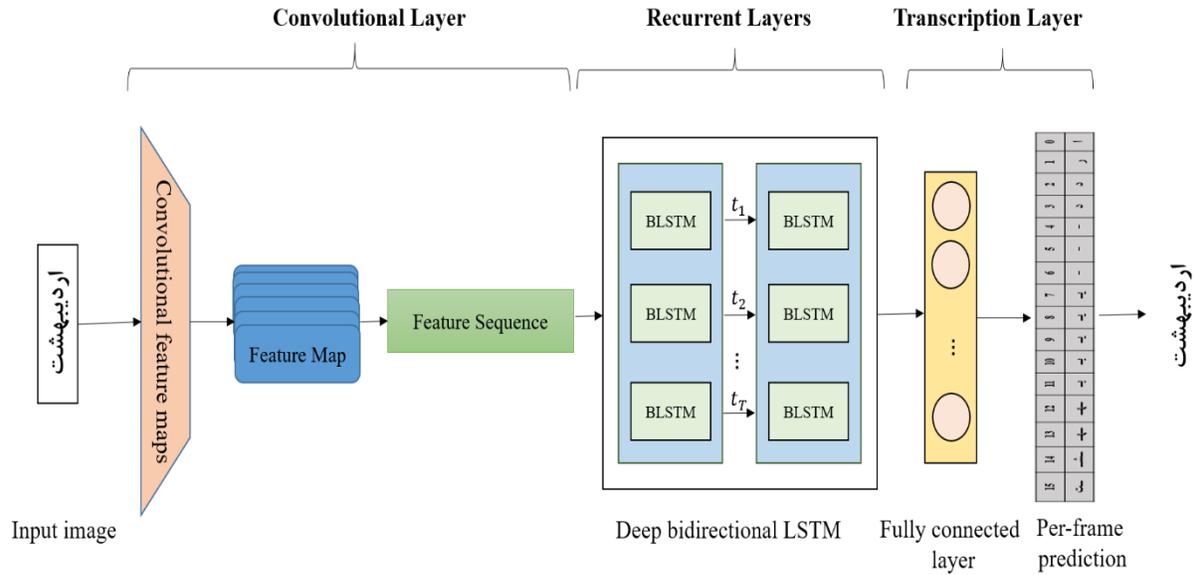

Fig. 8 The CRNN network consists of three modules: (1) Convolutional layers for feature extraction. (2) Recurrent layers consisting of Bidirectional LSTM layers to transform the sequence of features into predictions in each frame and (3) the Transcription layer enables the final output of labels based on repeated predictions in each frame.

### 4.1.2 Vision Transformer Architecture

The Vision Transformer (ViT) is a novel and innovative deep-based architecture tailored for computer vision tasks. ViT adapts the transformer model, originally tailored for natural processing, to the realm of image analysis [64]. What sets ViT apart is its reliance on self-attention mechanisms, enabling the model to examine the significance of various segments within the input image. Unlike conventional CRNN-based networks, ViT breaks down the image into non-overlapping patches, treating each as a "token" akin to words in language processing. To address the absence of inherent spatial information captured by CNNs, positional encoding is introduced to signify the location of each patch. Notably, in some applications, such as OCR, ViT employs solely the encoder component of the transformer architecture, as it utilizes the self-attention mechanisms critical for feature extraction and data relationships. Moreover, parallel processing in the ViT architecture, achieved through patch-level parallelism and multi-head self-attention, offers notable benefits, including improved training time and hardware efficiency, effective capture of long-range dependencies, scalability for different input image sizes, and a simplified model architecture. The proposed ViT architecture is depicted in Fig.9. Initially, the input image is partitioned into patches, which are subsequently converted into 1D vector embeddings. Each of these embeddings is accompanied by a learnable patch embedding and a positional encoding. The architecture is trained in an end-to-end manner to predict character sequences, using predefined tokens like [Go] to mark the sentence beginning and [s] to indicate spaces or the end of a character sequence.

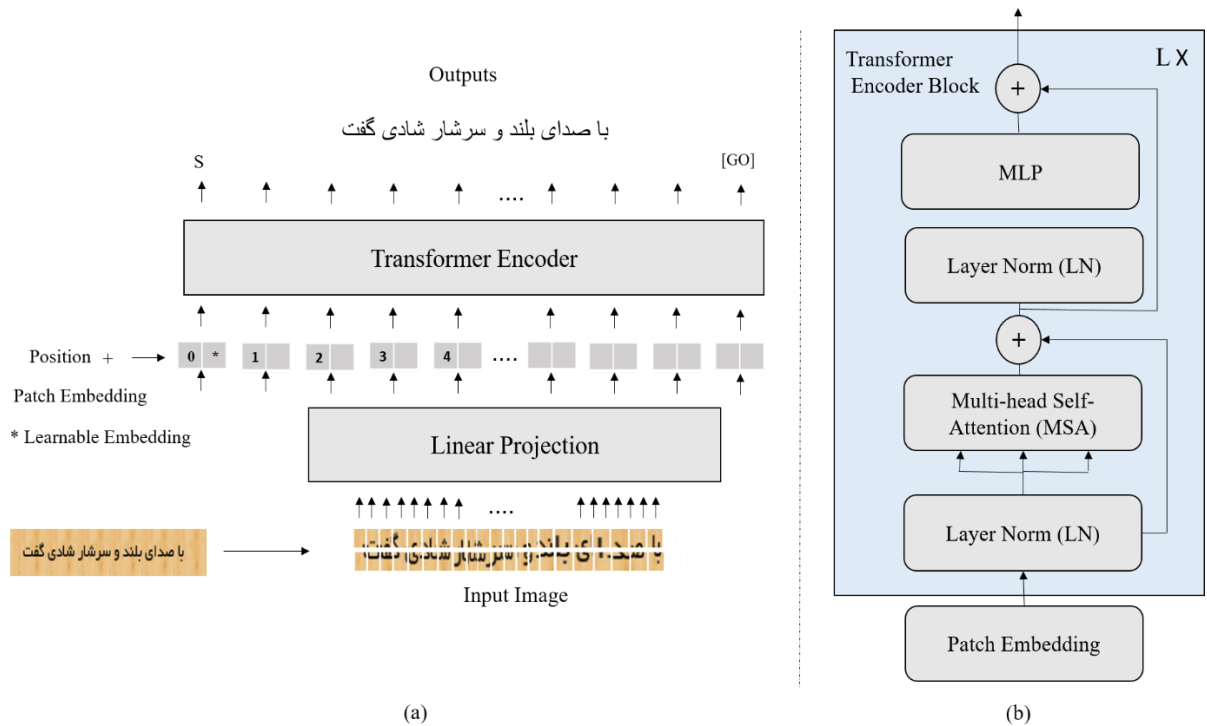

Fig. 9 (a) ViT architecture overview (b) The encoder module

## 4.2 EVALUATION METRICS

In the context of an OCR system, evaluation metrics are pivotal for ensuring the accurate conversion of textual information into machine-readable formats. Moreover, they provide a quantitative measure of how well the OCR system performs in correctly recognizing the characters, as well as identifying errors, missing characters, or misinterpretations. In light of this, the authors employed two well-known criteria, namely accuracy and the normalized Levenstein distance (NED), to evaluate the efficiency of the models [65]. The accuracy rate is determined by the number of completely correct words (CWrds) divided by the total word count (AllWds). In contrast, the NED provides a more sophisticated accuracy metric. It measures the dissimilarity between two strings, namely the recognized text (String1) and the ground truth (String2), by calculating the minimal number of edit operations (including insertions, deletions, and substitutions) needed to convert one string into the other.

The accuracy and the normalized edit distance are defined by the following equations:

$$Accuracy = \frac{CWrds}{AllWrds} * 100 \tag{1}$$

$$Normalized\ Levenstein\ Distance\ (NED) = \left(1 - \frac{Insertions + Deletions + Substitutions}{\max(Len(String1).Len(String2))}\right) \tag{2}$$

## 4.3 IMPLEMENTATION DETAILS

Both models undergo training utilizing mixed-precision on a machine equipped with a single Nvidia RTX 3070 (8GB of VRAM) GPU. The training process extends over 100 epochs, with a consistent batch size set at 128. The learning rate differs for each model while both models utilize the Adadelta optimizer with L1 regularization. Furthermore, the model is validated at intervals of every 1000 training iterations. Moreover, to address the potential issue of overfitting and improving training efficiency, an early stopping mechanism with a patience of 10 epochs is considered.

For model training, the dataset is divided into training, testing, and validation sets, maintaining a ratio of 0.75:0.15:0.15, respectively. Additionally, a maximum sequence length of S=40 is defined, and a charset of size S=153 is employed, incorporating Farsi numeric, character symbols, and punctuation marks. Conversely, to handle image preprocessing, a series of augmentation operations, such as rotation, scaling, and flipping, are applied to enhance the model's robustness and generalization. The remaining configurations for both models are maintained as discussed in [20, 62]. The training hyperparameters through the models are listed in Table 5.

### 4.4 RESULTS

This section presents the evaluation results of the newly proposed dataset, IDPL-PFOD, using two deep-based models: a convolutional recurrent neural network (CRNN) and a Transformer-based (ViT). a detailed exploration of the models is proposed in section 4.1. Moreover, to fairly comparison these two models, both of them are evaluated using the accuracy and the NED score which are explained in section 4.2.

Table 5 Training hyperparameters.

|  | Architecture | |
| --- | --- | --- |
| **Hyperparameter** | Vision Transformer (ViT) | CRNN-based |
| **Learning rate** | 1.0 [62] | 1.0e*10-3 |
| **Batch size** | 128 | 128 |
| **Regularization** | L1 | L1 |
| **Optimizer** | Adadelta | Adadelta |
| **Epochs** | 100 | 100 |

### 4.4.1 Qualitative Analysis

Figs. 10 and 11 illustrate the overall progress of the experiments on the IDPL-PFOD2 dataset, showcasing both accuracy and NED score throughout the training and validation stages for the CNN-based and the Vision Transformer architectures.

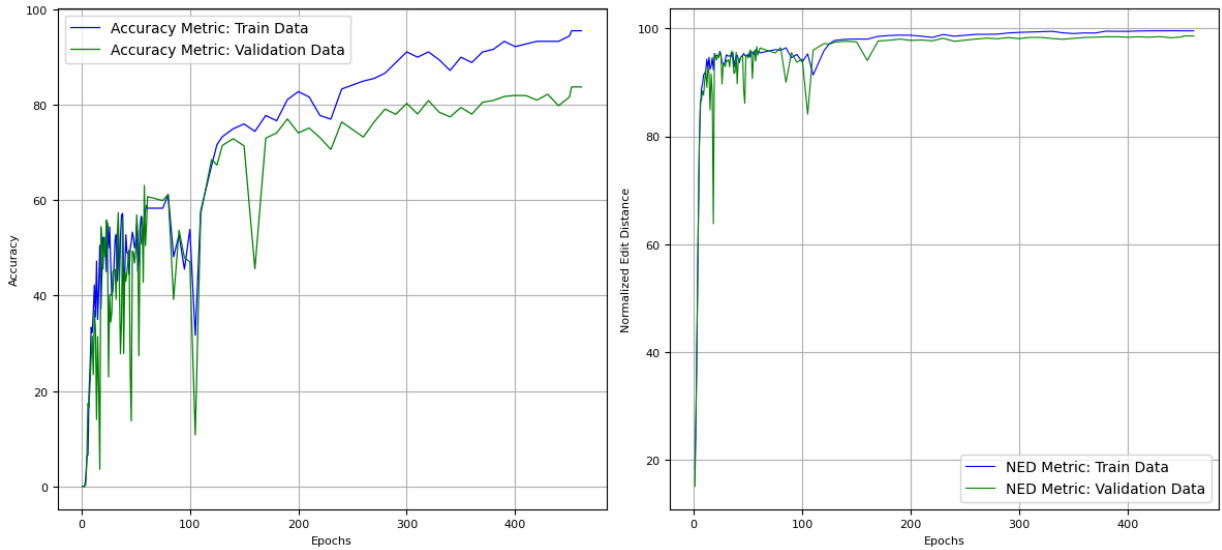

Fig. 10 evaluation results on the CRNN-based architecture (a) accuracy and (b) NED metrics.

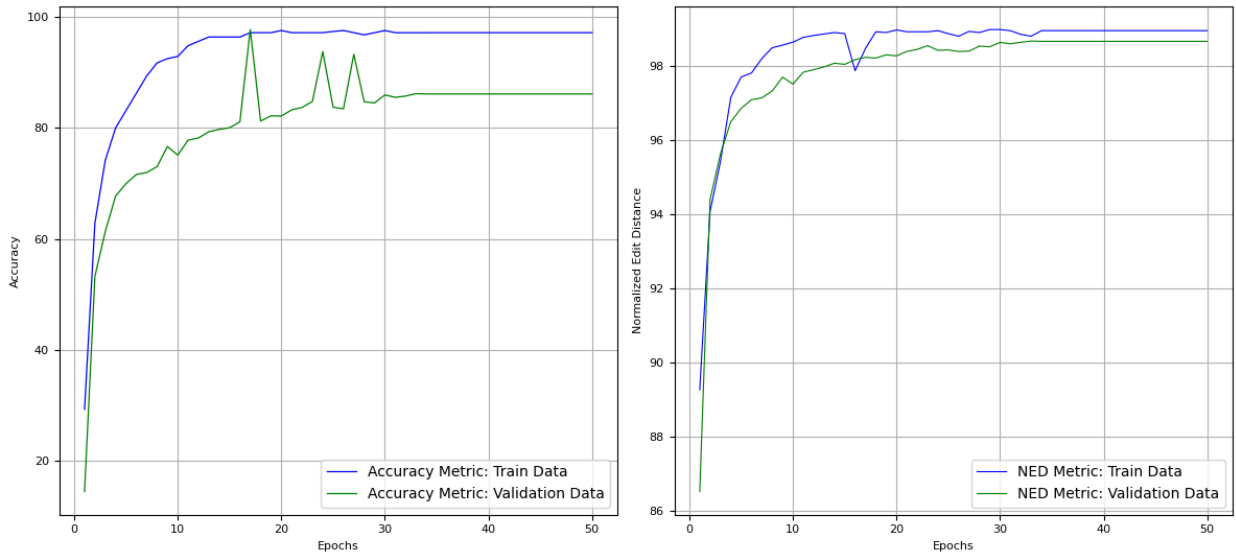

Fig. 11 evaluation results on the Vision Transformer architecture (a) accuracy and (b) NED metrics.

### 4.4.2 Quantitative Analysis

The ViT architecture outperforms the CRNN-based architecture with notable margins, exhibiting a superiority of 2.93% and 1.02% in absolute accuracy and NED scores, respectively, on the test dataset. Notably, both models were trained for 600 epochs, yet the transformer-based model demonstrated faster convergence, reaching optimal results around 40 epochs. In contrast, the CRNN-based model required the approximately 450 epochs for convergence. This disparity underscores the efficiency of transformer-based models, attributed to their non-reliance on sequential processing and inherent parallelization capability.

Table 6 provides a concise overview of the outcomes achieved by our proposed models. The CRNN-based model attains an accuracy of 78.49% and a normalized Levenshtein distance of 98.145%. In comparison, the ViT model surpasses these metrics with an accuracy of 81.32% and a NED score of 98.14%. These results highlight the commendable performance of both models, with the Vision Transformer exhibiting a slightly higher accuracy and comparable NED score, underscoring its efficacy in Farsi printed text recognition.

Table 6 results on CRNN-based and ViT models based on accuracy and NED metrics.

| Models | CRNN-based | | ViT | |
|---|---|---|---|---|
| metric | Accuracy (%) | NED score (%) | Accuracy (%) | NED score (%) |
| Train | 84.43 | 98.67 | 86.33 | 98.91 |
| Validation | 78.79 | 97.84 | 81.65 | 98.18 |
| Test | 78.49 | 97.72 | 81.32 | 98.14 |

## 5. Conclusions

Farsi language stands as a prominent and official language with millions of speakers worldwide and shares linguistic and structural similarities with other languages in the region such as Arabic and Urdu. The obstacles such as Farsi characters are written in a cursive manner and formidable similarities and character styles have hindered the development of an efficient and accurate Farsi OCR system. In addition to this, while most state-of-the-art deep-based architectures in the literature require substantial training samples to perform effectively, efforts to develop standard and publicly available large-scale datasets hold paramount importance in the OCR domain. Motivated by the aforementioned, this paper aims to propose a novel large-scale Farsi printed text dataset, called IDPL-PFOD2, comprising over 2 million image samples in various fonts, backgrounds, and noises.

To assess the suitability of the proposed dataset for the Farsi OCR task, the experimental process is carried over two state-of-the-art OCR models, namely a CRNN-based and a Vision Transformer architecture. To enable a fair comparison setting between these models, two well-known evaluation metrics such as accuracy and normalized edit distance were employed. The evaluation results demonstrate an accuracy rate of 78.49% and a normalized edit distance of 97.72% for the CRNN-based model, while the Vision Transformer architecture attains an accuracy of 81.32% and a normalized edit distance of 98.74%. As a consequence, these results highlight the dataset's effectiveness in addressing the complexities of Farsi printed text recognition and coins a solid benchmark for future research and inspiration for further work on this exciting and important research field.